\documentclass[lettersize,journal]{IEEEtran}
\usepackage{amsmath,amsfonts}
\usepackage{algorithmic}
\usepackage{algorithm}
\usepackage{array}
\usepackage[caption=false,font=normalsize,labelfont=sf,textfont=sf]{subfig}
\usepackage{textcomp}
\usepackage{stfloats}
\usepackage{url}
\usepackage{verbatim}
\usepackage{graphicx}
\usepackage{xcolor}%
\usepackage{cite}
\usepackage{multirow}

\begin{document}

\title{AM-MTEEG: Multi-task EEG classification based on impulsive associative memory}

\author{Junyan Li, Bin Hu*, Zhi-Hong Guan
\thanks{This work was supported in part by the National Natural Science Foundation of China under Fund 62322311. \textit{(*Corresponding author: B. Hu)}.}
\thanks{J. Li and B.~Hu are with the School of Future Technology, South China University of Technology, and also with the Pazhou Lab, Guangzhou 510006, China. (E-mails: 202164690091@mail.scut.edu.cn; huu@scut.edu.cn)} 
\thanks{Z.-H.~Guan is with the School of Artificial Intelligence and Automation, Huazhong University of Science and Technology, Wuhan 430074, China. (E-mail: zhguan@mail.hust.edu.cn)}
}

\markboth{IEEE Transactions on Biomedical Engineering,~Vol.~X, No.~X, 2024}%
{Shell \MakeLowercase{\textit{et al.}}: A Sample Article Using IEEEtran.cls for IEEE Journals}


\maketitle

\begin{abstract}
Electroencephalogram-based brain-computer interface (BCI) has potential applications in various fields, but their development is hindered by limited data and significant cross-individual variability. Inspired by the principles of learning and memory in the human hippocampus, we propose a multi-task (MT) classification model, called AM-MTEEG, which combines learning-based impulsive neural representations with bidirectional associative memory (AM) for cross-individual BCI classification tasks. The model treats the EEG classification of each individual as an independent task and facilitates feature sharing across individuals. Our model consists of an impulsive neural population coupled with a convolutional encoder-decoder to extract shared features and a bidirectional associative memory matrix to map features to class. Experimental results in two BCI competition datasets show that our model improves average accuracy compared to state-of-the-art models and reduces performance variance across individuals, and the waveforms reconstructed by the bidirectional associative memory provide interpretability for the model's classification results. The neuronal firing patterns in our model are highly coordinated, similarly to the neural coding of hippocampal neurons, indicating that our model has biological similarities.
\end{abstract}


\begin{figure*}[!t]
        \centering
        \includegraphics[width=0.9\linewidth]{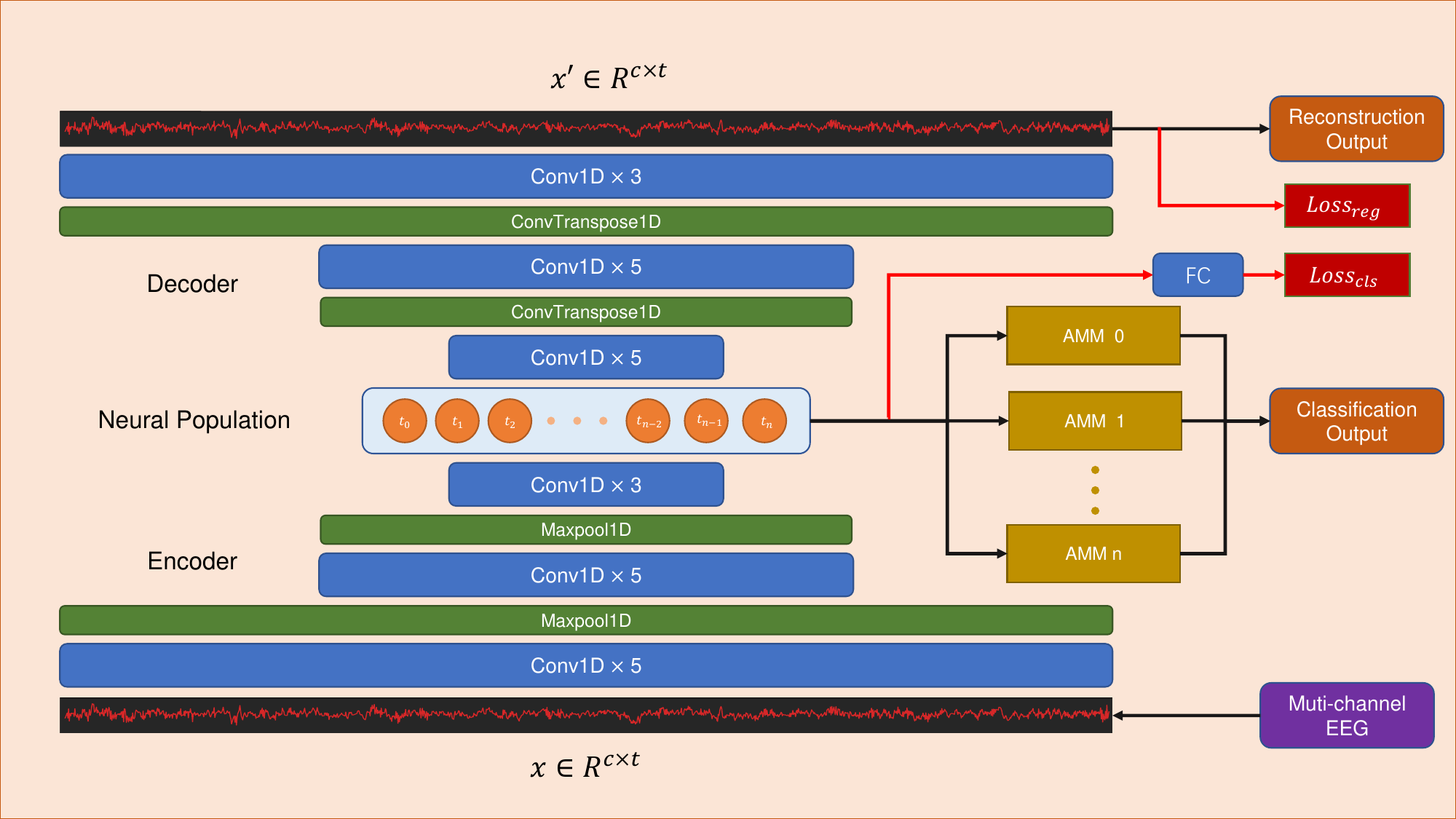}
        \caption{The overall architecture of the associative memory multi-task EEG (AM-MTEEG) model: Included are a convolutional encoder, impulsive neural population, a bidirectional associative memory (BAM) module, and a decoder with transposed convolution. Both the convolution layer and the transposed convolution layer use a convolution kernel of length 5 and are activated using the Relu function. In the experiment, we used 200  Leaky Integrate-and-Fire neurons as the impulsive neuron population. The encoder, impulsive neural population and decoder are trained by backpropagation (BP) with joint loss, and the BAM module is trained by Hebbian learning.}
        \label{fig:model}
    \end{figure*}

\section{Introduction}
    \IEEEPARstart{T}{he} brain-computer interface (BCI) can be defined as a system that translates a user’s brain activity patterns into messages or commands for interactive applications\cite{BCI_review}. The brain-computer interface effectively exchanges information between the brain and physical devices and has broad application prospects in medical rehabilitation and neuroscience research\cite{BCI_importance}. Most of the current BCI data comes from neural electrical signals recorded by electroencephalogram (EEG), which enables researchers to measure and decode human brain activity. A classic BCI paradigm, such as motor imagery (MI), basically consists of five parts: EEG acquisition, EEG preprocessing, feature extraction, classification, and task execution\cite{BCI_paradigm}. The key step is extracting the features of EEG signals and classifying them. Although EEG data collection technology has become highly mature, the classification of EEG data is still constrained by the following problems.
    \begin{itemize}
        \item Since EEG data has large variability and the representation of neural activity changes over time\cite{stable_BCI_manifold}, EEG classification models trained on different sample data are difficult to generalize to other samples.
        \item Most EEG classification models have low cross-individual classification accuracy\cite{EEGNet}, therefore models are only trained and tested on a single subject’s data\cite{eeg_10challenges}, resulting in insufficient data for training the model.
        \item Because BCI experiments are time-consuming\cite{EEG_emotion}, and are limited by the energy and time of the subjects, the amount of EEG data collected by a single subject is relatively small.
    \end{itemize}
    
    Although data-driven deep learning (DL) has achieved remarkable achievements in image recognition, natural language processing, and other fields, the above problems limit its application in EEG signal classification. On the other hand, due to ethical and safety considerations, the medical field has placed higher requirements on the interpretability of deep learning models\cite{DL_explain}. However, most current deep learning models have poor interpretability, which limits their application in BCI.

    \par To address the problem of large variability in EEG data, we were inspired by Zheng et al.\cite{MTL_EEG_tensor} and used multi-task learning (MTL) to model the EEG classification of each subject as a task, extract common features between samples of different subjects to achieve cross-individual training, and map the features to categories of each individual. In addition, to address the problem of less EEG data and poor interpretability of deep learning models, we hybridized the deep learning model with the associative memory model, which is an abstract model inspired by the memory principles of the real human brain. In contrast to deep learning, the human brain can learn effective features and make accurate mappings from very few demonstrations\cite{sensory_learning}. The ability of fast and few-shot learning comes from associative memory. The processes of learning and memory in the human brain can be categorized into three distinct phases\cite{cognitive_neuroscience}: encoding, storage, and retrieval. The brain structures responsible for encoding and retrieval have evolved over millions of years of species development and are refined through years of individual learning. In contrast, the neural architecture associated with the storage phase is established over a much shorter period. In this work, we propose replacing the brain's encoding mechanism with a deep learning encoder-decoder model trained across various samples. This model incorporates a layer of impulsive (also spiking) neurons to encode the necessary neural signals for associative memory formation. For each individual, we substitute the brain's storage and retrieval functions with a bidirectional associative memory (BAM) matrix. This matrix enables the mapping between two modalities: impulsive signals and classification categories. Consequently, the original EEG signals can be reconstructed by decoding the impulsive signals associated with specific category labels, thereby enhancing the interpretability of the model's classification process. Our major contributions can be summarized as follows.
    \begin{itemize}
        \item To address the challenges posed by significant individual variability and limited data in EEG analysis, we propose a model that integrates deep learning-based impulsive encoding with a Hebbian learning bidirectional associative memory network for EEG data classification, particularly in the context of brain-computer interfaces. The encoder in our model is designed to capture shared features across different samples, while the associative memory network effectively captures the variability among individual data. This work represents the application of impulsive neural networks and bidirectional associative memory to multi-task learning in EEG signal processing.
        \item For any motor imagery EEG signals, the impulsive neurons in our model exhibit specific firing patterns characterized by high synchrony. This synchrony suggests that neuronal activity is governed by a low-dimensional latent manifold, a feature consistent with the neural coding mechanisms observed in hippocampal neurons. This alignment indicates the bio-inspired nature of our model, reflecting its biological plausibility.
        \item Our model achieved an average accuracy of 86\% on the BCI Competition IV IIa dataset, surpassing existing state-of-the-art (SOTA) models and exhibiting minimal performance variance across different samples. Furthermore, the model can reconstruct EEG signals through the encoder-decoder framework. The EEG signals reconstructed by the bidirectional associative memory network can be compared with the event-related potentials (ERPs) of each category, thereby enhancing the interpretability of the model.
    \end{itemize}

\section{Related Work and Study Motivation}
    In the classification of EEG data, particularly within the field of BCI applications, a common approach involves using common spatial patterns (CSP) for feature extraction\cite{CSP}, followed by classification algorithms such as LDA and SVM. While CSP can significantly enhance EEG features, traditional machine learning models struggle to recognize more complex EEG patterns. Consequently, deep learning models have demonstrated greater flexibility in EEG classification for complex BCI tasks. For example, Lawhern et al.\cite{EEGNet} proposed EEGNet, which applies separable two-dimensional convolutions to EEG classification problems. Liu et al.\cite{FBMSNet} combine the same spatiotemporal convolution with filter banks and propose FBMSNet, which mixes deep convolution to extract temporal features at multiple scales and then performs spatial filtering to mitigate volume conduction. Similarly, Altaheri et al.\cite{ATCNet} introduced ATCNet, a convolutional neural network with temporal attention mechanisms for EEG classification. This model achieved an average accuracy of 85.4\% on the BCI Competition IV IIa dataset, setting a new state-of-the-art performance on this dataset.

    \par Multi-task learning optimizes multiple loss functions simultaneously, allowing different tasks to share the same features. Compared to single-task learning models, MTL leverages more data from different tasks\cite{MTL_survey}, enabling the learning of more generalized representations. Moreover, MTL addresses the challenges of high individual variability and limited sample size in EEG data. Zheng et al.\cite{MTL_EEG_tensor} developed an effective algorithm where each subject's sample is treated as a separate task, utilizing regularized tensors. This method employs Fisher's discriminant criterion for feature selection and optimizes using the alternating direction method of multipliers (ADMMs). In addition to MTL, the use of ensemble learning can also significantly reduce the variability of EEG. For example, Yu et al.\cite{baysian} used a dynamic ensemble Bayesian filter to assemble models. The ensemble models can cope with variability in signals and improve the robustness of online control.

    \par Spiking neural networks (SNNs), inspired by biological neural systems, offer advantages such as low power consumption and high interpretability, making them widely applicable across various tasks. Diehl et al.\cite{stdp_mnist} implemented an SNN using unsupervised STDP learning for handwritten digit classification, achieving 95\% accuracy on the MNIST dataset. Xu et al.\cite{snn_emg} proposed a spiking convolutional neural network (SCNN) for electromyography (EMG) pattern recognition, which can be used in prosthesis control and human-computer interaction. The effectiveness of SNNs in multi-task learning has also been demonstrated. For instance, Cachi et al.\cite{TM_SNN} proposed TM-SNN, which uses different spiking thresholds to represent different tasks while sharing the same structure and parameters across tasks.

    \par Models that integrate Hebbian associative memory neural networks with deep learning have been shown to enhance performance across a range of tasks. Hu et al.\cite{AM_stable} proposed that spiking neural networks using Hebbian learning can provide stable and fault-tolerant associative memory. Miconi et al.\cite{BP_Hebb} combined Hebbian rule-based associative memory with traditional backpropagation neural networks, achieving efficient learning on small-sample image datasets. Building on this, Wu et al.\cite{global_local} applied a similar structure to spiking neural networks, where the network weights are updated through both global learning (backpropagation) and local learning (Hebbian learning). This hybrid approach has demonstrated remarkable performance in tasks such as fault-tolerant learning, few-shot learning, and continual learning.

    \par Despite advancements in BCI classification models utilizing neural manifolds and multi-task learning to reduce inter-individual variability, challenges remain in feature encoding clarity, interpretability, and consistent performance across individuals. These limitations hinder the widespread adoption of BCI technology in fields such as medical research and rehabilitation. To address these issues, our model integrates deep learning-based SNN with BAM networks. The BAM framework enhances the interpretability of BCI performance across different individuals, while the shared features extracted through multi-task learning improve classification stability across individuals.
    
\section{The AM-MTEEG Model}
    As illustrated in Fig.\ref{fig:model}, our proposed multi-task learning model consists of a spiking encoder and an associative memory classifier. The spiking encoder utilizes a one-dimensional convolutional neural network to extract signal features, which are then fed into a population of spiking neurons, encoding the input into low-dimensional spiking representations. A convolutional neural network decoder is employed to reconstruct the EEG signals. The associative memory classifier assigns an associative memory matrix to each classification task, mapping the encoded spikes to multi-task categories.

    \par The training of the multi-task learning model is divided into two phases. In the first phase, we combine self-supervised learning with label-guided training to optimize the encoder-decoder model. Multi-channel EEG signals are used simultaneously as both input and target, training the spiking encoder to reconstruct the one-dimensional EEG signals. In the second phase, we freeze the parameters of the spiking encoder and train the corresponding associative memory matrices for different tasks. Here, we represent the input and category labels as input-output pattern pairs $\{\mathbf{x}_i,\mathbf{y}_i\}$, where $\mathbf{x}_i\in R^{n\times t}$ is the low-dimensional spiking representation vector produced by the spiking encoder, with $n$ representing the number of neurons in the population and $t$ denoting the time series length after CNN encoding. $\mathbf{y}_i$ is the target vector where the labels are one-hot encoded. The associative memory matrix matches the input patterns to the corresponding output patterns by forming bidirectional hetero-associative memory.
    
    \subsection{Convolutional Feature Extractor}
        As shown in Fig.\ref{fig:model}, our convolutional feature extractor consists of an encoder $E$ and a decoder $D$ consisting of one-dimensional convolutions, where the convolution kernel length of the convolutional layer is 5 and the ReLU function is used for activation. Unlike the existing EEG classification models based on 2D convolution\cite{EEGNet,ATCNet}, to achieve EEG data classification while maintaining the structure of the original EEG signal as much as possible, we only used 1D convolution for feature extraction. Therefore, the convolution kernel parameters to be trained are reduced from $c\times n^2$ to $c\times n$, where $c$ is the number of signal channels and $n$ is the convolution kernel size, which allows us to use larger convolution kernels. In the motor imagery task, we used the CNN model architecture as shown in Table \ref{tab:model_architecture}. The input signal $\mathbf{x}\in R^{c\times t}$ is downsampled to 1/4 of the original length by two one-dimensional maximum pooling in the encoder to obtain the hidden signal
        \begin{equation}
            \mathbf{h} = E(\mathbf{x}), \mathbf{h}\in R^{n\times t/4}.
        \end{equation}
        The encoded signal is directly input into the spiking neuron as the current through the fully connected layer, recording the spike sequence $\mathbf{S}_p\in R^{n\times t/4}$ emitted by the neuron. In the decoder stage, the hidden layer spike are mapped by the fully connected layer and then upsampled to the original length by two one-dimensional deconvolutions.
        \begin{equation}
            \mathbf{x}' = D(\mathbf{S}_p), \mathbf{x}'\in R^{c\times t}.
        \end{equation}
        In this process, the Encoder-Decoder model and the impulsive neural population obtain a low-dimensional representation of EEG activity through autoregressive learning.

        \begin{table*}[!t]
        \caption{The size of each module output signal\label{tab:model_architecture}}
        \centering
        \renewcommand\arraystretch{1.2}
        \begin{tabular}{c|cccccc}
\hline
Blocks                            & Layer           & $N_{conv}$ & Size                                    & Stride & Activation & Output    \\
\hline
\multirow{6}{*}{Encoder}          & Input           &         & $(C, T)$                                  &        &            & $(C,T)$     \\
                                  & Conv1D Block    & 3       & 5                                       & 1      & ReLU       & $(128,T)$   \\
                                  & AvgPool1D       &         & 2                                       & 2      &            & $(128,T/2)$ \\
                                  & Conv1D Block    & 5       & 5                                       & 1      & ReLU       & $(256,T/2)$ \\
                                  & AvgPool1D       &         & 2                                       & 2      &            & ($256,T/4)$ \\
                                  & Conv1D Block    & 5       & 5                                       & 1      & ReLU       & $(256,T/4)$ \\
\hline
\multirow{3}{*}{Neual Population} & FC              &         & $(256,200)$                               &        &            & $(200,T/4)$ \\
                                  & LIF Neurons     &         & 200                                     &        &            & $(200,T/4)$ \\
                                  & FC              &         & $(200,256)$                               &        &            & $(256,T/4)$ \\
\hline
\multirow{5}{*}{Decoder}          & Conv1D Block    & 3       & 5                                       & 1      & ReLU       & $(128,T/4)$ \\
                                  & ConvTranspose1D &         & 8                                       & 2      &            & $(128,T/2)$ \\
                                  & Conv1D Block    & 5       & 5                                       & 1      & ReLU       & $(128,T/2)$ \\
                                  & ConvTranspose1D &         & 8                                       & 2      &            & $(128,T)$   \\
                                  & Conv1D Block    & 5       & 5                                       & 1      & ReLU       & $(C,T)$     \\
\hline
Associative Memory Classifier     & AMM             &         & $(200\times T/4,N_{class})$ &        &            & $N_{class}$ \\
\hline
\end{tabular}
    \end{table*}

    \subsection{Impulsive neural population}
        We use the encoder output signal as a current $\mathbf{I}$, input it into the Leaky Integrate-and-Fire (LIF) neuron population, and convert it into a discrete spike train $\mathbf{s}$. The change of LIF membrane potential $u$ with discrete time $t$ is
        \begin{equation}
            \tau\frac{du}{dt} = -u + RI(t),
        \end{equation}
        when the membrane potential is greater than $u_{th}$, the neuron generates a spike, and the membrane potential is set to 0. To facilitate computer simulation, we use its differential form
        \begin{equation}
        \label{eq:LIF}
            \begin{aligned}
                u^t &= (1-\tau)u^{t-1}-s^{t-1}u_{th}+\sum_{i}(w_iI_i^{t-1}),\\
                s^t &= step(u^t-u_{th}),
            \end{aligned}
        \end{equation}
        where $\tau$ is the decay constant, $w_i$ is the synaptic weight of the synapse $i$, $s^t \in \{0,1\}$ is the spike fired at $t$, and $I_i^{t-1}$ represents the input current of the synapse $i$ at time $t-1$. When the membrane potential is greater than $u_{th}$, the neuron generates a spike, and the membrane potential is set to 0 at the next time $t+1$. As shown in eq.\ref{eq:LIF}, This process uses the unit step function 
        $$step(x) = 
            \begin{cases}
                1&x\ge0, \\
                0&x<0.
            \end{cases}$$
        However, when training the encoder using backpropagation, calculating the gradient of the step function poses a challenge. Since the step function is discontinuous, its gradient results in an impulse response 
        $$\delta(x)=\begin{cases}
            +\infty&x=0,\\
            0&x\neq0.
        \end{cases}$$
        Therefore the gradient of membrane potential is
        \begin{equation}
            \begin{aligned}
                \nabla u^{t-1} &= \frac{\partial u^t}{\partial u^{t-1}}\nabla u^t + \frac{\partial s^{t-1}}{\partial u^{t-1}}\nabla s^{t-1} \\
                &= (1-\tau)\nabla u^t + \delta (u^t-u_{th})\nabla s^{t-1} \\
                &= \nabla u^t[(1-\tau)+u_{th}\delta (u^t-u_{th})].
            \end{aligned}
        \end{equation}
        where $\delta (u^t-u_{th})$ makes it difficult to train the encoder. As shown in Figure 2, we use the surrogate gradient method\cite{Surrogate_gradient} to replace the impulse function with a rectangular window function
        \begin{equation}
            rect(x) = 
            \begin{cases}
                1&|x|\le0.5 \\
                0&|x|>0.5
            \end{cases}.
        \end{equation}
        The membrane potential gradient using the surrogate gradient is
        \begin{equation}
            \begin{aligned}
                \nabla u^{t-1} &= \frac{\partial u^t}{\partial u^{t-1}}\nabla u^t + (\frac{\partial s^{t-1}}{\partial u^{t-1}})'\nabla s^{t-1} \\
                &= (1-\tau)\nabla u^t + rect(u^t-u_{th})\nabla s^{t-1} \\
                &= \nabla u^t[(1-\tau)+u_{th}rect(u^t-u_{th})].
            \end{aligned}
        \end{equation}

        \begin{figure}[!t]
        \centering
        \includegraphics[width=0.9\linewidth]{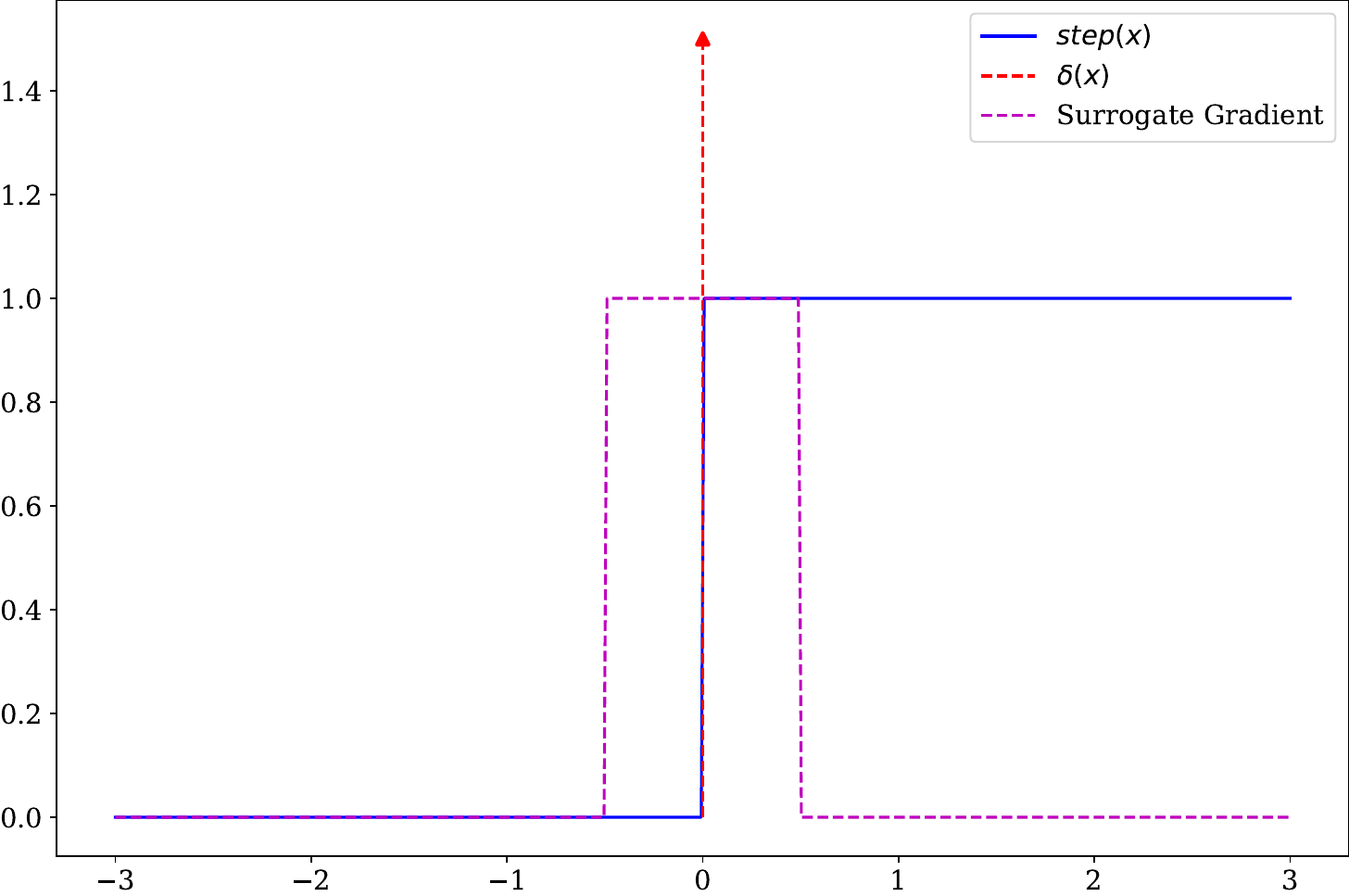}
        \caption{The step function, impulse function, and surrogate gradient functions}
        \label{fig: Surrogate Gradient}
        \end{figure}
    
        \par As illustrated in Fig.\ref{fig:model}, at each time step t, the signal processed by the convolutional neural network is treated as input current, which is passed through a fully connected layer into multiple LIF neurons in the hidden layer. The hidden layer generates spikes, and these spike sequences encode essential information from the original EEG signal. Subsequently, we employ a self-supervision approach, using a convolutional neural network decoder to reconstruct the EEG signal. To enhance the model's ability to pre-classify the input EEG while performing signal reconstruction, we connect the hidden layer neurons to an auxiliary classifier for preliminary classification. Here we use a trainable fully connected layer, and we decompose the training loss function into reconstruction loss $L_{reg}$ and classification loss $L_{cls}$, where the reconstruction loss is MSE, i.e. $L_{reg}(\mathbf{x},\mathbf{x'})=\frac{1}{n}\sum_i(x_i-x_i')^2$, and the classification loss uses cross-entropy loss, i.e. $L_{cls}(\mathbf{x},label) = -\sum_ilabel_ilog(x_i)$. The joint loss is
        \begin{equation}
            L = L_{reg} + \lambda L_{cls},
        \end{equation}
        where $\lambda$ is an artificially set mixing factor, which we set to 0.1 here, and $L_{cls}$ is calculated by the hidden layer spike through the auxiliary classifier.

    \subsection{Associative Memory Classifier}
        In order to perform efficient training and accurate classification on a physiological electrical signal dataset with a small amount of data, we use a bidirectional associative memory method to map the pulse activity to the label. Let the input-output pattern pair be $\{\mathbf{x}_k,\mathbf{y}_k\}$, where $\mathbf{x}_k\in R^n$ is the input column vector and $\mathbf{y}_k\in R^m$ is output one-hot vector. In the memory retrieval stage, the iterative process of the input-output pattern pair $\{\mathbf{x}_k,\mathbf{y}_k\}$ using the associative memory matrix (AMM) $\mathbf{W}_k$ is\cite{BAM_eq}
        \begin{equation}
        \label{eq:BAM}
            \begin{aligned}
                \mathbf{y}_k^{t+1} &= sgn(\mathbf{W}_k\mathbf{x}_k^t), \\
                \mathbf{x}_k^{t+1} &= sgn(\mathbf{W}_k^T\mathbf{y}_k^t),
            \end{aligned}
        \end{equation}
        where $sgn=\begin{cases}
            -1&x\le0 \\
            +1&x>0
        \end{cases}$. We write eq. \ref{eq:BAM} as the sum of each term,
        \begin{equation}
            \begin{aligned}
                y_{t+1}^i = \sum_{j=1}^nw^{ij}x_t^j, \\ 
                x_{t+1}^i = \sum_{j=1}^mw^{ji}y_t^j.
            \end{aligned}
        \end{equation}
        Its differential form is
        \begin{equation}
        \label{eq:BAM_diff}
        \begin{aligned}
            \frac{dy^i}{dt} = \sum_{j=1}^nw^{ij}x^j, \\ 
            \frac{dx^j}{dt} = \sum_{i=1}^mw^{ij}y^i.
        \end{aligned}
        \end{equation}
        
        \par When the associative memory system is stable, $y^{t+1}=y^t, x^{t+1}=x^t$, so the above process is actually optimizing the energy function of the system\cite{BAM_energy}
        \begin{equation}
            E = -\mathbf{y}^T\mathbf{Wx}.
        \end{equation}
        We can write it as
        \begin{equation}
        \label{eq:BAM_energy}
            E = \sum_i\sum_j -y^iw^{ij}x^j.
        \end{equation}
        Therefore $\frac{\partial E}{\partial w^{ij}} = -y^ix^j$, and when $w^{ij}=sgn(y^ix^j)$, the energy function is minimum. During the training stage, the associative memory matrix $\mathbf{W}_k$ of the task $k$ is
        \begin{equation}
            \mathbf{W}_k = \sum_j\mathbf{y}_k^j\mathbf{x}_k^{jT}.
        \end{equation}
        From the above, we can see that bidirectional associative memory is a type of Hebbian learning. When the pre and post-synaptic membranes emit spikes at the same time, the synaptic connection will be strengthened. This process is similar to the synaptic learning mechanism of hippocampal neurons\cite{hebbian_hippo}.

        \par Next, we will prove the convergence of the system. Let$\dot x^j, \dot y^i$ be the time derivatives of $x^j, y^i$. Then the rate of change of the energy function E is
        \begin{equation}
        \label{eq:diff_E}
            \dot E = \sum_j\frac{\partial E}{\partial x^j}\dot x^j + \sum_i\frac{\partial E}{\partial y^i}\dot y^i.
        \end{equation}
        From eq. \ref{eq:BAM_energy}, we know
        \begin{equation}
        \label{eq:derive_E}
            \begin{aligned}
                \frac{\partial E}{\partial x^j} = \sum_i -y^iw^{ij}, \\
                \frac{\partial E}{\partial y^i} = \sum_j -x^jw^{ij}.
            \end{aligned}
        \end{equation}
        Substitute eq.\ref{eq:derive_E} and eq. \ref{eq:BAM_diff} into eq. \ref{eq:diff_E}, we can get
        \begin{equation}
            \dot E = -\sum_j(\sum_iw^{ij}y^i)^2 - \sum_i(\sum_jw^{ij}x^j)^2,
        \end{equation}
        where
        \begin{equation}
            \begin{aligned}
                (\sum_iw^{ij}y^i)^2\ge0,\\
                (\sum_jw^{ij}x^j)^2\ge0.
            \end{aligned}
        \end{equation}
        Therefore, $\dot E\le0$ always holds true, and the dynamic changes of the system will cause $E$ to continue to decrease. Considering the use of the $sgn$ function, the system will gradually converge to a stable value.

        \par Since we use one-hot encoded $\mathbf{y}^j$ as the output pattern, after applying the $sgn$ function, only the maximum value is set to 1, while the remaining values are set to -1. This system stabilizes after a single iteration. Therefore, during the testing phase, for a given task sample $\mathbf{x}_i$, the classification result is obtained using the associative memory matrix
        \begin{equation}
            label^i = \mathop{\arg\min}\limits_{i}(\mathbf{W}^i\mathbf{x}^i).
        \end{equation}
        Considering all the pattern pairs $\{\mathbf{x}_k, \mathbf{y}_k\}$ to be stored, for any input $\mathbf{x}_i$ in the prediction phase
        $$
        \mathbf{y}_i = \sum_k\mathbf{y}_k\mathbf{x}_k^T\mathbf{x}_i,
        $$
         this process is equivalent to taking the cosine similarity between the current input $\mathbf{x}_i$ and $\mathbf{x}_k$ in all pattern pairs as the average output $\mathbf{y}_k$ of the weighted calculation.

\section{Experimental Results}
    We applied our model to the classic motor imagery BCI paradigm, which classifies users' EEG signals into different motor actions based on EEG classification models. In this study, we utilized the BCI Competition dataset and achieved an average accuracy of 94\% and 86\% on two of its subsets, respectively.
    \subsection{Experimental datasets}
        \paragraph{BCI Competition III Iva\cite{BCI_III_IVa}} This dataset is a binary classification dataset, including right-hand and foot movement imagery tasks performed by 5 subjects. Each task includes 118 channels of EEG signals obtained at a sampling rate of 100 Hz within 3 seconds.
        \paragraph{BCI Competition IV IIa\cite{BCI_IV_2a}} This dataset is a four-category dataset, including motor imagery tasks of the left hand, right hand, feet, and tongue performed by 9 subjects. Each task includes 22 channels of EEG signals and 3 channels of EOG signals obtained at a sampling rate of 250 Hz within 3 seconds.

    \subsection{Performance evaluation}
        \paragraph{Comparative studies}
        As shown in Table \ref{tab:bci_IV_IIa} and Table \ref{tab:bci_III_IVa}, we compared our proposed model with other models. Compared to existing SOTA models, our model achieved comparable accuracy and surpassed the current SOTA model in terms of average accuracy on the BCI Competition IV IIa dataset. In contrast to other multi-task models, our proposed model exhibited the smallest standard deviation in accuracy across different individuals, indicating its ability to provide stable classification performance across individuals. Additionally, when extending to new tasks, our model only requires retraining the associative memory matrix, and the Hebbian-like learning used in this process is highly efficient, demonstrating good scalability.

        \begin{table*}[!t]
        \caption{BCI Competition IV IIa comparison experiment accuracy\label{tab:bci_IV_IIa}}
        \centering
        \begin{tabular}{c|ccccc}
        \hline
        Tasks & Zheng et al.\cite{MTL_EEG_tensor} & DMTL-BCI\cite{DMTL_BCI} & EEGNet (Non-MTL)\cite{EEGNet} & ATCNet (Non-MTL, SOTA)\cite{ATCNet} & Ours\\
        \hline
        1 & 0.840 & 0.835 & 0.858 & \textbf{0.885} & 0.810 \\
        2 & 0.573 & 0.490 & 0.615 & 0.705 & \textbf{0.793} \\
        3 & 0.549 & 0.927 & 0.886 & \textbf{0.976} & 0.879 \\
        4 & \textbf{0.959} & 0.670 & 0.749 & 0.810 & 0.831 \\
        5 & 0.912 & 0.713 & 0.559 & 0.830 & \textbf{0.948} \\
        6 & 0.826 & 0.637 & 0.521 & 0.736 & \textbf{0.897} \\
        7 & 0.792 & 0.808 & 0.896 & \textbf{0.931} & 0.862 \\
        8 & 0.835 & 0.800 & 0.833 & \textbf{0.903} & 0.879 \\
        9 & 0.819 & 0.817 & 0.795 & \textbf{0.910} & 0.844 \\
        \hline
        AVG & 0.790 & 0.753 & 0.745 & 0.854 & \textbf{0.860} \\
        STD & 0.131 & 0.120 & 0.139 & 0.086 & \textbf{0.045} \\
        \hline
        \end{tabular}
        \end{table*}
        
        \begin{table*}[!t]
        \caption{BCI Competition III Iva comparison experiment accuracy\label{tab:bci_III_IVa}}
        \centering
        \begin{tabular}{c|ccccc}
        \hline
        Tasks & EEGNet (Non-MTL)\cite{EEGNet} & EDPNet(Non-MTL)\cite{edpnet} &STL-Overlap\cite{STL_overlap} & Zheng et al.\cite{MTL_EEG_tensor} & Ours\\
        \hline
        aa & \textbf{1.000} & \textbf{1.000} & 0.857 & 0.911 & 0.958\\
        al & 0.688 & 0.884 & 0.982 & \textbf{1.000} & 0.975\\
        av & 0.582 & 0.704 & 0.643 & 0.768 & \textbf{0.838}\\
        aw & 0.795 & 0.835 & 0.964 & \textbf{1.000} & 0.975\\
        ay & 0.516 & 0.679 & 0.911 & 0.929 & \textbf{0.950}\\
        \hline
        AVG & 0.716 & 0.820 &0.871 & 0.921 & \textbf{0.942}\\
        STD & 0.176 & 0.118 &0.122 & 0.084 & \textbf{0.052}\\
        \hline
        \end{tabular}
        \end{table*}
        
\begin{figure}[!t]
        \centering
        \includegraphics[width=0.9\linewidth]{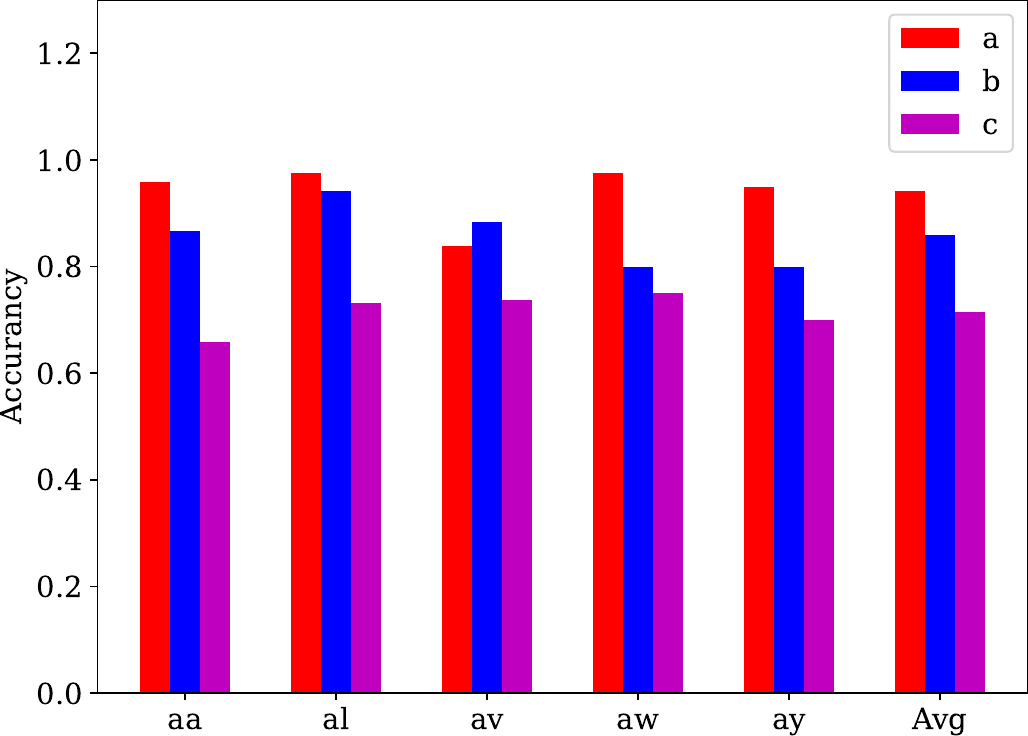}
        \caption{Ablation experiments on the BCI Competition III Iva binary classification dataset. (a) Original model (b) Model with spiking neurons removed (c) Model with a fully connected network using gradient descent instead of the associative memory classifier. The results show that the spiking neurons and bidirectional associative memory networks included in the model can improve classification performance.}
        \label{fig: ablation}
        \end{figure}

        \paragraph{Ablation studies}
        We evaluated our proposed model on the BCI Competition III Iva dataset and compared it against two alternative models: (a) a model where spiking neurons were replaced with the tanh function, and (b) a model where the associative memory matrix was replaced by a fully connected layer trained via gradient descent. The results, as shown in Fig.\ref{fig: ablation}, demonstrate that the original model outperformed the simplified models in terms of accuracy on most samples. These findings suggest that both the spiking computation and the bidirectional associative memory classifier used in our model contribute to improved performance.

        \begin{figure*}[!t]
        \centering
        \subfloat[]{\includegraphics[width=0.4\linewidth]{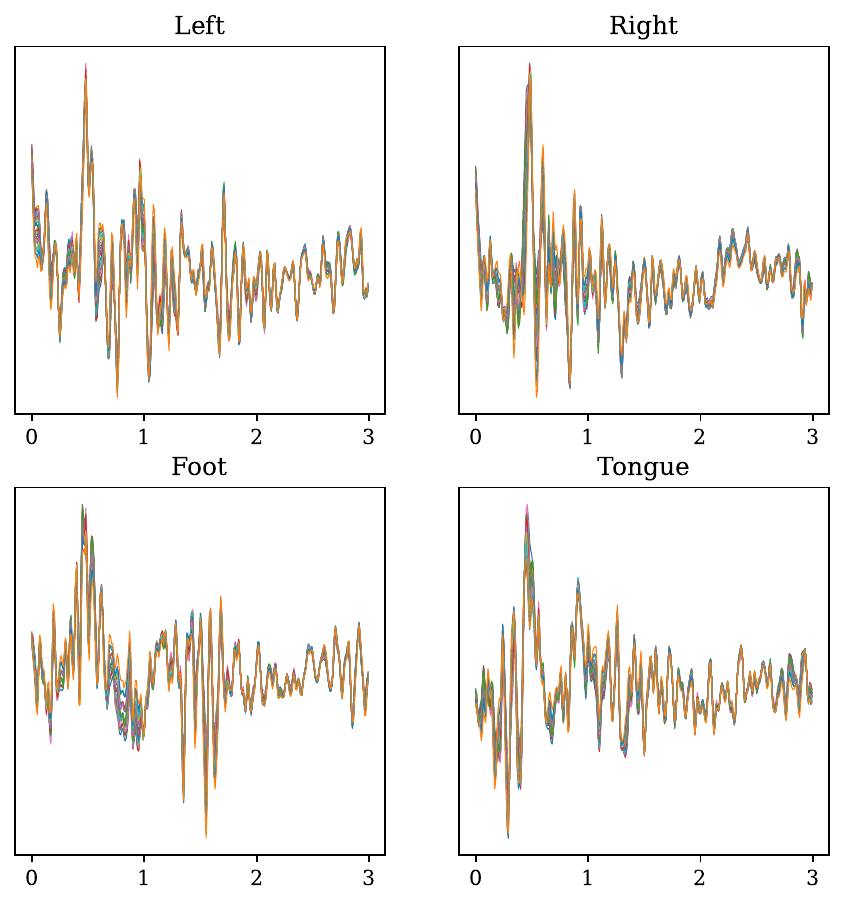}
        \label{fig:waveform}}
        \hfil
        \subfloat[]{\includegraphics[width=0.4\linewidth]{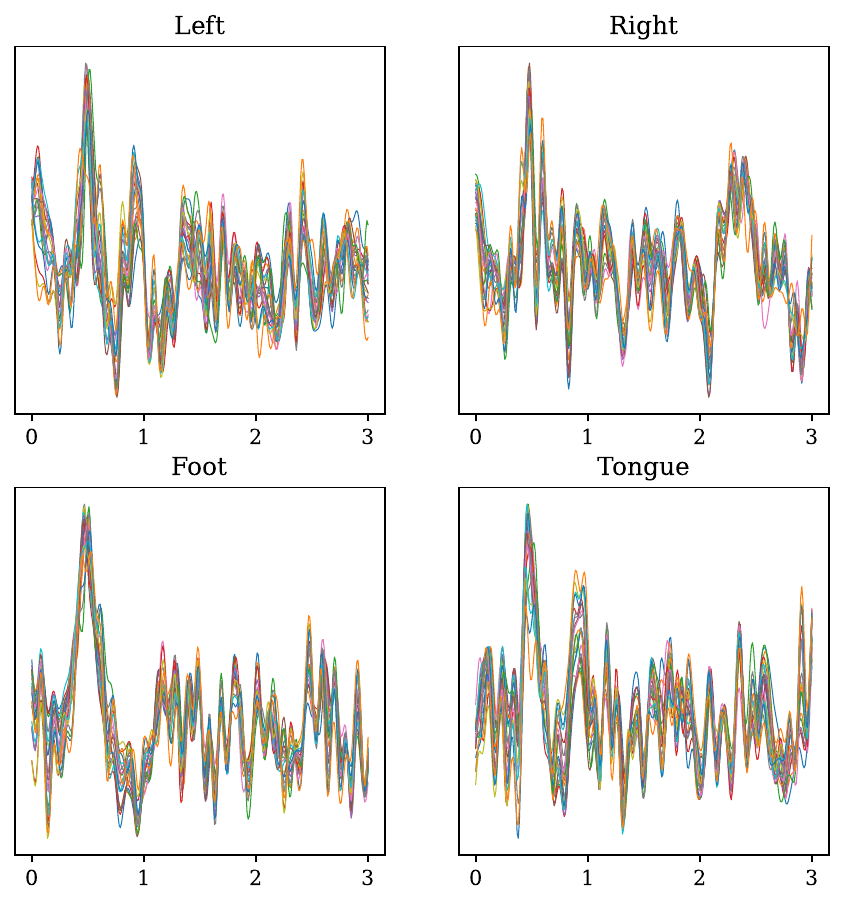}%
        \label{fig:ERP}}
        \caption{(a) Reconstructed waveform and (b) Event-related potential of each movement in BCI Competition IV IIa}
        \label{fig:ERP&wave}
        \end{figure*}

    \subsection{Model interpretability}
\begin{figure*}[!t]
        \centering
        \includegraphics[width=0.79\linewidth]{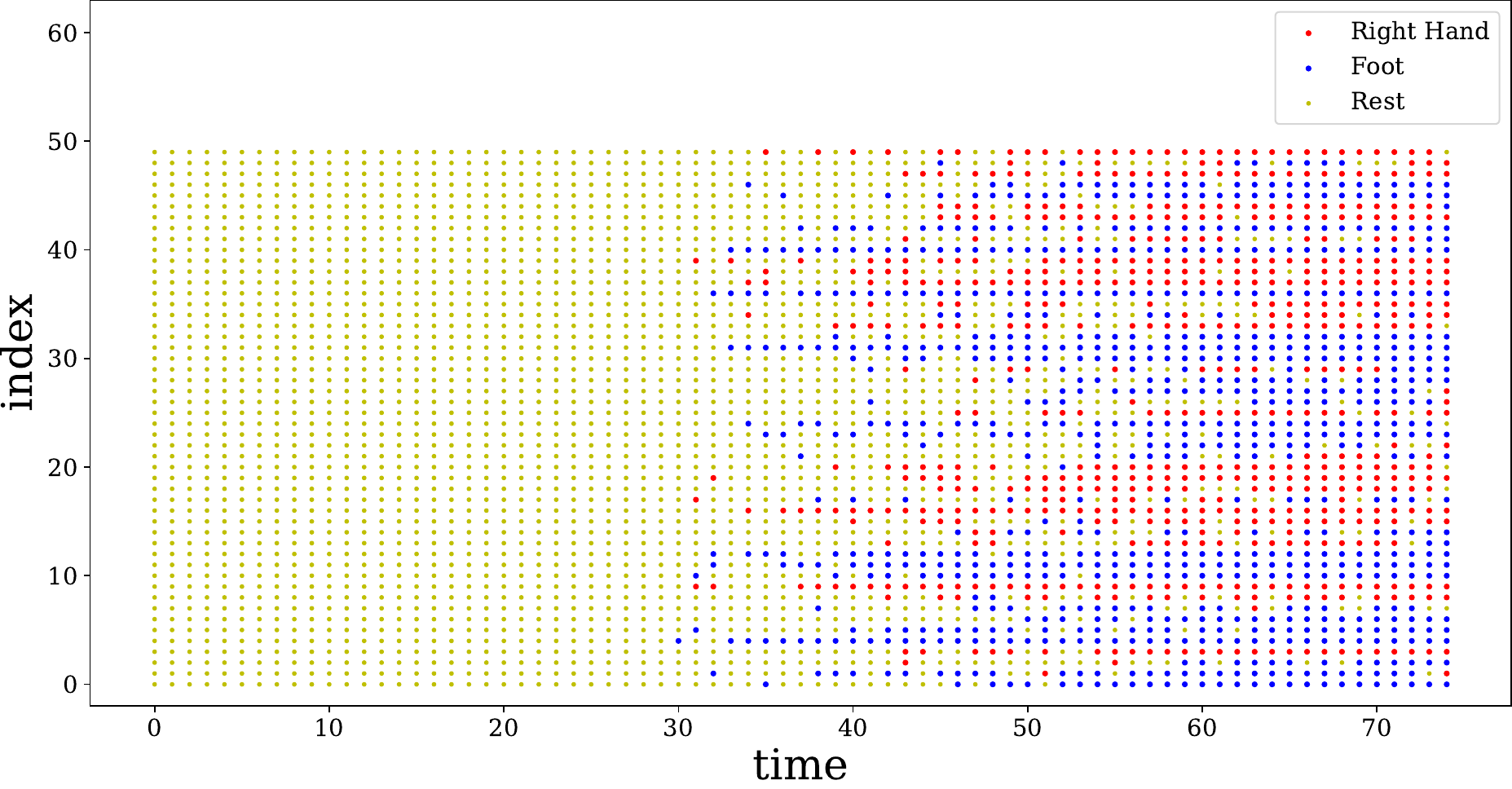}
        \caption{The time of neuronal spikes associated with two types of movement}
        \label{fig: spike}
        \end{figure*}

        Due to the reversibility of bidirectional associative memory, we input the classification label into the associative memory matrix to obtain the characteristic pulse sequence corresponding to any category
        \begin{equation}
            x_i^{label} = sgn(\mathbf{W}_i^T\mathbf{y}^{label}).
        \end{equation}
        The spiking sequences obtained through the bidirectional associative memory on the BCI Competition III Iva dataset are shown in Fig. \ref{fig: spike}. It can be observed that the spikes from most individual neurons represent a fixed category, and these spikes exhibit a high degree of synchrony. This suggests that the spiking neurons in the hidden layer share similarities with the neural population coding observed in the hippocampus of the human brain\cite{hippo_manifold}.

        \par We then used the decoder to reconstruct the original EEG data corresponding to the labels in the BCI Competition IV IIa dataset, obtaining characteristic waveforms for the four motor imagery categories. As shown in Fig.\ref{fig:waveform}, the reconstructed EEG signals reveal distinct waveforms for each of the four categories. When comparing these waveforms with the actual ERP from the real data, as illustrated in Fig.\ref{fig:ERP}, we observe a similarity between the reconstructed waveforms and the ERP. The greater the similarity between these two waveforms, the higher the confidence in the model's correct classification.

\section{Discussion}
    Current research on multi-task learning for EEG classification is limited, but its effectiveness in cross-subject classification on highly variable EEG signals has been demonstrated. Our proposed method enables rapid adaptation to new individuals once the encoder has been trained, a capability derived from the bidirectional associative memory network. For brain-computer interface applications, our model can be fitted to new individuals using only a small number of samples, which facilitates the interaction of BCI devices across different users.

    \par With the advent of neuromorphic computing, specialized chips designed for spiking neural networks have been developed\cite{loihi,tianjic}. Compared to CPUs and GPUs based on the Von Neumann architecture, neuromorphic circuits can be implemented using in-memory computing digital circuits or memristor-based analog circuits. These circuits offer advantages such as high parallel efficiency, low energy consumption, and compact size. The computation of spiking neural populations on such neuromorphic circuits can be employed in edge scenarios, promoting the domestic and miniaturized use of BCI devices.
    
    \par Although initially designed to address single-task cross-individual EEG classification, our multi-task learning model can also be applied to single-individual multi-task or multi-individual multi-task scenarios when tasks are correlated. Therefore, our model holds promise for achieving multimodal EEG decoding.
    
\section{Conclusion}
    This article has developed AM-MTEEG, a multi-task EEG classification model based on impulsive associative memory. This model effectively integrates impulsive neural representations from deep learning with bidirectional associative memory networks to address challenges in the brain-computer interface field, such as limited EEG data, high variability, and the lack of interpretability in deep learning models. Through the multi-task learning framework, our model treats each subject's classification task as an independent task and leverages cross-subject training to extract shared features and facilitate feature sharing across individuals. Our model demonstrated superior performance on the BCI Competition dataset, achieving accuracy that surpasses existing state-of-the-art models on the BCI Competition IV IIa dataset, while minimizing classification performance variance across different samples. These results validate our model's effectiveness in extracting common EEG features, capturing data variability, and handling cross-subject classification tasks. Future work will focus on further integrating associative memory networks with deep impulsive neural networks, as well as exploring the joint learning of Hebbian rules and gradient descent.




 
%

\bibliographystyle{IEEEtran}


 




\end{document}